%% file: cloudMatch.tex
\documentclass[10pt,twocolumn,letterpaper]{article}
\usepackage[utf8]{inputenc}
\usepackage[T1]{fontenc}
\usepackage{iccv}              


\definecolor{iccvblue}{rgb}{0.21,0.49,0.74}
\usepackage[pagebackref,breaklinks,colorlinks,allcolors=iccvblue]{hyperref}

\usepackage{caption}
\usepackage{subcaption}
\usepackage{graphicx}
\usepackage[american]{babel}
\usepackage{microtype}
\usepackage{multirow}
\usepackage{booktabs}
\usepackage{algorithm}
\usepackage{algorithmicx}
\usepackage{algpseudocode}

\usepackage{bm}

\usepackage{wrapfig}
\usepackage{pifont}
\usepackage{multirow}
\usepackage[export]{adjustbox}
\usepackage{color}
\usepackage{colortbl}
\usepackage{lipsum}
\usepackage{pifont}
\usepackage{bbding}
\usepackage[dvipsnames]{xcolor}
\usepackage{array}

\usepackage[capitalize]{cleveref}
\usepackage{float}
\crefname{section}{Sec.}{Secs.}
\Crefname{section}{Section}{Sections}
\Crefname{table}{Table}{Tables}
\crefname{table}{Tab.}{Tabs.}
\definecolor{red}{RGB}{255,0,0}
\definecolor{blue}{RGB}{0,0,255}
\definecolor{green}{RGB}{0,255,0}
\definecolor{mygray}{gray}{.9}
\definecolor{mygray2}{gray}{.5}
\definecolor{mywarning}{RGB}{233,144,61}
\definecolor{mygreen}{RGB}{93,174,86}
\definecolor{codefunc}{RGB}{73,122,234}

\usepackage[american]{babel}
\usepackage{microtype}

\definecolor{mygreen}{RGB}{0,154,85}
\definecolor{myy}{RGB}{126,95,0}
\definecolor{myred}{RGB}{212,121,116}
\definecolor{myblue}{RGB}{184, 134, 73}
\definecolor{mynewgreen}{RGB}{113,188,169}
\definecolor{mypurple}{RGB}{123,104,238}
\colorlet{R1}{myblue}
\colorlet{R2}{mypurple}
\colorlet{R3}{myred}
\colorlet{R6}{mypurple}
\definecolor{mycite}{RGB}{73,123,184}
\colorlet{cite}{mycite}

\def\paperID{1234} 
\def\confName{ICCV}
\def\confYear{2030}

\title{CloudMatch: Weak-to-Strong Consistency Learning for Semi-Supervised Cloud Detection}

\author{
	 Jiayi Zhao$^1$, Changlu Chen$^2$, Jingsheng Li$^1$, Tianxiang Xue$^1$, and Kun Zhan$^{1,\star}$\\
	1. School of Information Science and Engineering, Lanzhou University\\
	2. Faculty of Data Science, City University of Macau, Macau, China\\
	{\small \url{https://github.com/kunzhan/CloudMatch}}
}

\begin{document}
\maketitle
\input{body}

{
\small
\bibliographystyle{ieeenat_fullname}
\bibliography{tgrs}
}
\end{document}

%% file: body.tex
\begin{abstract}
	Due to the high cost of annotating accurate pixel-level labels, semi-supervised learning has emerged as a promising approach for cloud detection. In this paper, we propose CloudMatch, a semi-supervised framework that effectively leverages unlabeled remote sensing imagery through view-consistency learning combined with scene-mixing augmentations. An observation behind CloudMatch is that cloud patterns exhibit structural diversity and contextual variability across different scenes and within the same scene category. Our key insight is that enforcing prediction consistency across diversely augmented views, incorporating both inter-scene and intra-scene mixing, enables the model to capture the structural diversity and contextual richness of cloud patterns. Specifically, CloudMatch generates one weakly augmented view along with two complementary strongly augmented views for each unlabeled image: one integrates inter-scene patches to simulate contextual variety, while the other employs intra-scene mixing to preserve semantic coherence. This approach guides pseudolabel generation and enhances generalization. Extensive experiments show that CloudMatch achieves good performance, demonstrating its capability to utilize unlabeled data efficiently and advance semi-supervised cloud detection.
\end{abstract}
\section{Introduction}
Cloud detection is a fundamental task in the remote sensing domain, aiming to accurately identify and localize cloud regions from satellite imagery. It plays a critical role in a wide range of downstream applications, including land cover classification~\cite{rs13224708}, agricultural monitoring~\cite{WEISS2020111402}, and climate modeling~\cite{li2016remote}, where the presence of clouds can significantly influence the Earth observation data.

Supervised cloud detection methods~\cite{cdnet,dabnet,hrcloudnet} are effective in cloud detection tasks, as they can identify and locate cloud regions based on existing labeled data. However, these methods heavily rely on large-scale annotated datasets with precise pixel-level labels. Generating such annotated data is time-consuming and labor-intensive, and this issue becomes more prominent when faced with the massive data volume and high resolution of modern satellite images. Thus, semi-supervised learning has emerged as a highly promising solution, which utilizes both labeled data and massive unlabeled data for training.

In semi-supervised learning, consistency regularization has become a predominant paradigm~\cite{Abuduweili_2021_CVPR,NEURIPS2020_44feb009,NEURIPS2019_1cd138d0}.
The core idea is to enforce the model to produce consistent predictions across various augmented views. By minimizing the discrepancy between predictions from these diverse views, consistency regularization enhances model robustness and generalization to unseen data. It has been successfully adopted in cloud detection methods. For example, SSCDnet~\cite{sscd} generates high-confidence pseudolabels using a dual-threshold dynamic selection strategy combined with output-level domain adaptation, and MTCSNet~\cite{rs15082040} introduces a cross-supervision framework to alleviate prediction inconsistencies caused by model initialization.

However, a central challenge lies in designing effective augmented views for consistency learning. To achieve reliable consistency, it is crucial to adopt augmentation strategies that preserve semantic integrity.
Existing strategies\cite{cutmix,DBLP:classmix,mixup} enrich data diversity by blending regions from different samples. Yet in pixel-level segmentation tasks like cloud detection, such blending often introduces semantic ambiguity and intra-view inconsistency, which can confuse the model and degrade performance, especially in visually subtle scenarios.

To address this challenge, we propose an inter- and intra-scene mixing augmentation approach for semi-supervised cloud detection. This method effectively leverages intra-image semantic consistency and inter-image sample diversity to enhance the robustness and generalization capability of the model. Specifically, inter-scene mixing augmentation enhances data diversity by blending regions from multiple images. This strategy leverages complementary information from different scenes to enrich the semantic content of training samples, while avoiding excessive reliance on external data sources. However, it also introduces structural inconsistency between the newly generated samples and the original real samples.
Complementarily, intra-scene mixing augmentation operates within a single image, where different regions are independently subjected to weak and strong augmentations before being combined. This process not only generates diverse training samples but also preserves the global structural consistency of the original image, thereby improving the model's ability to adapt to local variations.
By integrating both intra- and inter-scene augmentation mechanisms into a unified framework, our approach substantially enhances the diversity and representativeness of the training data. As a result, the model achieves superior performance in challenging scenarios.

With such a variety of weakly and strongly augmented views, we first introduce a weak-to-strong pseudo supervision loss. Beyond pseudo supervision, we propose a weak-to-strong view-consistency loss specifically designed for semi-supervised cloud detection. This loss enforces consistency across augmented views by implicitly aligning class-wise output distributions~\cite{xu2024structure,xia2025hierarchical}, effectively realizing the core goal of view-consistency learning: maximizing the correlation between different views of the same instance. As a result, it encourages more discriminative and stable feature representations even under limited annotations.

To this end, we propose CloudMatch, a unified semi-supervised framework that fully exploits unlabeled remote sensing images through consistency-driven learning. CloudMatch comprises two key components: (1) a dual strong augmentation module that combines inter-scene mixing (patches from different scenes) and intra-scene mixing (within-category variations) to support weak-to-strong pseudo supervision; and (2) a weak-to-strong view-consistency loss that aligns weakly and strongly augmented views at the class level, enhancing representation robustness. These augmentation strategies are carefully designed to capture the structural diversity and contextual variability inherent in real-world cloud imagery. The synergy between augmentation and consistency learning enables CloudMatch to achieve superior cloud detection performance even under limited annotated data.

A key distinction of CloudMatch is that inter- and intra-scene mixing are not treated as independent augmentations, but are explicitly embedded into the consistency learning framework. Furthermore, weak-to-strong view-consistency is enforced on these mixed views. Inter- and intra-scene mixing play complementary roles in CloudMatch. Intra-scene mixing preserves semantic coherence while exposing structural variations of cloud patterns within the same scene, making it suitable for reliable pseudo-label supervision. In contrast, inter-scene mixing introduces broader contextual diversity caused by changes in surface reflectance, terrain, and acquisition conditions.

In summary,  the contributions of this study are as follows:
\begin{itemize}
\item We design a view consistency loss that aligns weakly and strongly augmented views at the class level, encouraging semantically consistent predictions and enhancing representation robustness under limited annotations.

\item We propose a dual-path augmentation module that generates diverse and complementary views through both inter-scene mixing (cross-scene patch blending) and intra-scene mixing (within-category transformations), supporting effective consistency regularization by promoting both inter-view and intra-view interaction.

\item We reconfigure the Biome dataset for semi-supervised cloud detection, and demonstrate through extensive experiments that CloudMatch consistently outperforms strong baselines across multiple benchmarks.
\end{itemize}
\section{Related Work}
\subsection{Semi-Supervised Segmentation}
Semi-supervised image segmentation has progressed rapidly in recent years, aiming to alleviate the dependence on large-scale pixel-level annotations by jointly exploiting a small set of labeled data and a large pool of unlabeled data. The key challenge lies in effectively mining useful supervision from unlabeled samples to improve model generalization and segmentation accuracy.

Existing semi-supervised segmentation methods can be broadly divided into three categories: self-training, pseudolabeling, and consistency regularization. Self-training methods iteratively refine the model by generating pseudolabels from an initial teacher network and using them to train a student network. For example, Xie et al.\cite{Xie_2020_CVPR} demonstrated that a teacher-student pipeline can substantially boost segmentation performance by expanding the training set with pseudo-annotated images.
pseudolabeling methods emphasize the reliability of generated labels, as noisy pseudolabels can degrade performance. SoftMatch\cite{softmatch}, for instance, maintains a balance between pseudolabel quantity and quality, ensuring that the model benefits from both abundant and accurate supervision. TrustMatch\cite{He_2025_ICCV} integrates bias-aware pseudolabel refinement with interpretable trust evaluation, explicitly quantifying the bias tendency of each pseudolabel through a composite score, thereby adaptively suppressing misleading supervision signals and achieving superior generalization.
Consistency regularization further enhances performance by encouraging stable predictions across different augmentations of the same input. UniMatch~\cite{Yang_2023_CVPR} extends this principle through a dual-stream perturbation strategy, where two strongly augmented views are aligned with a shared weak view, leading to improved consistency and robustness. Subsequently, UniMatch-v2\cite{10839097} integrates the feature-level and input-level augmentations of UniMatch into a single learnable stream, and introduces Complementary Dropout to fully exploit dual-stream training. RankMatch\cite{Mai_2024_CVPR} selects a set of representative reference pixels through orthogonal selection as agents, and by modeling the relationships among agents, ensures that the agent-level correlations between weakly and strongly augmented views remain consistent in terms of ranking probability distributions.

Although these approaches achieve remarkable results in general vision tasks, directly applying them to remote sensing cloud detection remains challenging. This is due to complex background interference, spectral similarity between clouds and bright surfaces, and diverse cloud morphology. These challenges motivate the development of tailored semi-supervised strategies for cloud detection, as discussed in the following section.
\subsection{Semi-Supervised Segmentation for Cloud Detection}
Semi-supervised learning has shown remarkable potential in cloud detection, primarily because generating precise annotations for remote sensing images is both costly and labor-intensive, particularly in complex regions where cloud boundaries are ambiguous.
At the same time, a large number of unlabeled cloud images are readily available from satellites, providing a rich source of data for SSL techniques. For example, Guo et al.~\cite{9570362} introduced an unsupervised domain adaptation framework that transfers trained cloud detection models to new satellite platforms without requiring additional annotations, highlighting that the primary challenge lies in the scarcity of labeled data. As a result, SSL approaches have become an active research direction for cloud detection tasks.

Recent semi-supervised cloud detection methods mainly improve pseudolabel reliability, introduce consistency or cross-supervision constraints, or integrate auxiliary strategies to better leverage unlabeled data.

SSCDnet~\cite{sscd} employs a dual-threshold pseudolabel strategy to obtain reliable pseudolabels,  effectively mitigating the interference of noisy labels during self-training and enhancing model performance. Additionally,  they introduce feature-level and output-level domain adaptation techniques to reduce the domain distribution discrepancy between labeled and unlabeled images, thereby improving the prediction accuracy of SSL networks. SSAL-CD~\cite{10155420} combines semi-supervised learning with active learning, utilizing a small number of labeled images and a large number of unlabeled images to jointly train deep neural networks for pixel-level cloud detection. This framework enhances consistency through mutual supervision between two segmentation networks,  while active learning selects the most valuable samples for annotation. In-extensive Nets~\cite{doi:10.1080/15481603.2022.2147298} adopts a cross-supervision paradigm, where two base networks are jointly trained by combining supervised learning on labeled data with mutual supervision on unlabeled data. Each network leverages the other's predictions as additional supervision signals, effectively reducing label noise and improving model robustness. MTCSNet~\cite{rs15082040} employs a teacher-student cross-supervision framework enhanced by near-infrared band inputs and robust data augmentations. CrossMatch\cite{10769516} uses the pseudolabels of weakly augmented data from one view to supervise the model training in another view, and maximizes the dissimilarity of feature representations across views to ensure that complementary information provides more valuable guidance for the model training in the other view.
U-MCL\cite{10891590} generates a patch-wise uncertainty map for each unlabeled image and adaptively adjusts the mask ratio for pseudolabel denoising accordingly. Meanwhile, this uncertainty map is also used to model masked unlabeled images for inferring unseen regions.
MUCA\cite{11062866} introduces a multiscale uncertainty consistency regularization and a cross-teacher-student attention mechanism to guide the student network in constructing more discriminative feature representations through complementary features from the teacher network.

These methods not only introduces more prior information but also achieves consistency constraints across different batches of the same image and intra-batch accuracy constraints, further enhancing the accuracy and robustness of cloud detection and remote sensing image segmentation.

Despite these advances, most existing SSL cloud detection methods either rely heavily on the quality of pixel-wise pseudolabels or impose consistency at the prediction level without explicitly aligning cross-view semantics at a global category level. In contrast, CloudMatch introduces a view-consistency loss that aligns weak and strong augmented views at the semantic category level, coupled with a dual-scene (intra- and inter-scene) mixing strategy that expands feature diversity while preserving structural coherence, thereby yielding stronger generalization under limited annotations.
\section{CloudMatch for Semi-Supervised Cloud Detection}
We present CloudMatch, a semi-supervised framework specifically designed for cloud detection. The proposed method effectively leverages both limited labeled data and abundant unlabeled samples through two key components: (1) a hybrid scene-mixing augmentation that integrates intra-scene and inter-scene mixing strategies, and (2) a view-consistency learning scheme that enforces prediction consistency across differently augmented views. We first introduce the problem formulation and the overall CloudMatch, followed by detailed explanations of its supervisions.
\subsection{CloudMatch}
\begin{figure*}[!t]
\centering
\includegraphics[width=0.98\textwidth]{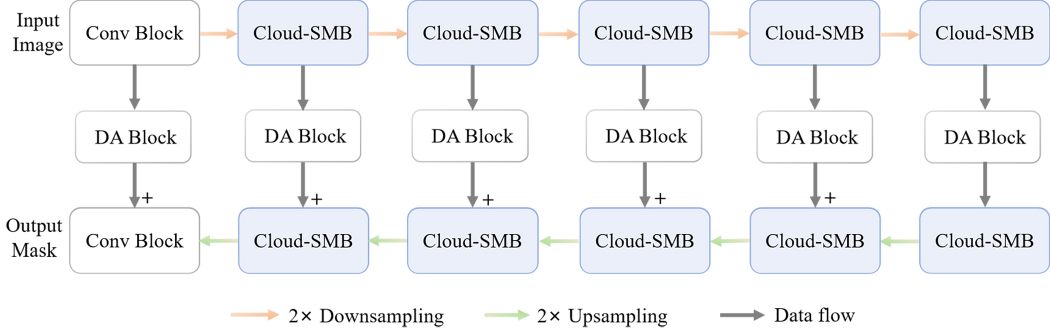}
\caption{CD-Mamba network architecture. The model is based on a U-shaped structure, integrating convolutional modules with Cloud-SMB (Cloud Spatial Mamba Block) modules and incorporating dual-attention blocks (DA Blocks) in the skip connections to enhance cloud boundary detection accuracy.}
\label{CD-Mamba}
\end{figure*}
\begin{figure*}[!t]
\centering
\includegraphics[width=0.98\textwidth]{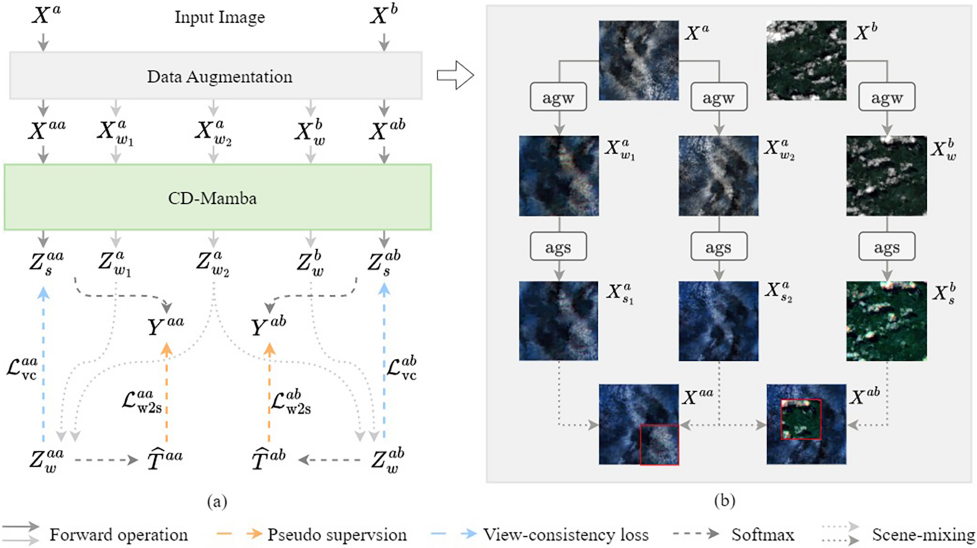}
\caption{CloudMatch architecture and scene-mixing augmentation framework. The prediction represents a probabilistic prediction map, where each pixel value ranges from $[0,1]$, while the pseudolabel corresponds to a binary prediction map with values of $\{0,1\}$.}\label{architecture2}
\end{figure*}
CloudMatch is designed to fully exploit the abundant unlabeled remote sensing data to enhance cloud detection performance under limited annotations. The training process involves two parallel learning streams: one supervised with labeled data and the other unsupervised with unlabeled data. Formally, let $\mathcal{D}_l = \{(\bm{X}_i, \bm{T}_i)\}_{i=1}^{n_l}$ denote the labeled dataset and $\mathcal{D}_u =\{\bm{X}_i\}_{i=1}^{n_u}$ the unlabeled dataset, where $\bm{X}_i$ and $\bm{T}_i$ represent the input image and its corresponding ground-truth mask, respectively. We argue that most existing semi-supervised methods primarily adopt convolutional networks as segmentation backbones. However, their inherent limitation to local feature modeling constrains the performance of semi-supervised approaches on remote sensing datasets. To address this issue, we adopt CD-Mamba~\cite{xueJARS2025}, a cloud detection network based on long-range dependency modeling with Mamba, as the backbone of our approach, whose overall architecture is illustrated in Figure~\ref{CD-Mamba}. CD-Mamba integrates convolutional operations with Mamba-based state-space modeling into a unified and lightweight cloud detection network, enabling simultaneous pixel-level interactions and long-term patch-wise dependency modeling.

For a labeled sample $(\bm{X}, \bm{T})$, the labeled image $\bm{X}$ is fed into CD-Mamba~\cite{xueJARS2025}, to extract latent features $\bm{Z}$. A softmax layer is then applied to obtain the prediction map $\bm{Y}$, which is optimized using a standard cross-entropy loss with respect to the ground truth $\bm{T}$.

For unlabeled data, as illustrated in Figure~\ref{architecture2}, two samples $\bm{X}^a, \bm{X}^b \in \mathcal{D}_u$ are randomly selected and subjected to both weak and strong augmentations. Specifically, weakly augmented views $\bm{X}_{w_1}^a, \bm{X}_{w_2}^a$, and $\bm{X}_{w}^b$ and strongly augmented views $\bm{X}_{s_1}^a, \bm{X}_{s_2}^a$, and $\bm{X}_{s}^b$ are generated. The strong views are then combined through intra-scene and inter-scene mixing operations to produce two composite augmented images, denoted as $\bm{X}^{aa}$ and $\bm{X}^{ab}$. These images, $\bm{X}^{aa}, \bm{X}^{ab},\bm{X}_{w_1}^a, \bm{X}_{w_2}^a$, and $\bm{X}_{w}^b$, are then fed into the CD-Mamba backbone to obtain corresponding predictions. Finally, weak-to-strong pseudo-supervision is applied to enforce prediction consistency between weakly and strongly augmented views.

For unlabeled data, pseudolabels are used to guide the learning process. Weak augmentations are first applied to $\bm{X}^a$ and $\bm{X}^b$ to obtain feature representations $\bm{Z}_w^a$ and $\bm{Z}_w^b$. Using the same intra- and inter-scene mixing strategy, these features are combined into $\bm{Z}_w^{aa}$ and $\bm{Z}_w^{ab}$, from which hard pseudolabels $\hat{\bm{T}}^{aa}$ and $\hat{\bm{T}}^{ab}$ are derived. These pseudolabels supervise the predictions of the corresponding strongly augmented views $\bm{Y}^{aa}$ and $\bm{Y}^{ab}$ through cross-entropy loss.
To further enhance consistency, a weak-to-strong view-consistency loss aligns the weakly and strongly augmented features, from weak views $\bm{Z}_w^{aa}, \bm{Z}_w^{ab}$ to strong views $\bm{Z}_s^{aa}, \bm{Z}_s^{ab}$.

As illustrated in Figure~\ref{architecture2}(b), two unlabeled remote sensing images, $\bm{X}^a$ and $\bm{X}^b$, are randomly sampled from different scenes in the unlabeled dataset.
To construct diverse yet semantically coherent views, we adopt a scene-mixing augmentation pipeline consisting of weak and strong transformations, denoted as ${\rm agw}(\cdot)$ and ${\rm ags}(\cdot)$, respectively. Weak augmentation ${\rm agw}(\cdot)$ is limited to basic spatial transformations (random resizing, cropping, and flipping), while strong augmentation ${\rm ags}(\cdot)$ builds upon weak augmentation to perform more impactful random enhancements, including color jittering, grayscale conversion, Gaussian blur, and CutMix region mixing.

Weak augmentations are first applied to generate multiple weak views:
\begin{align}
\bm X_{w_1}^a = {\rm agw}(\bm X^a), \quad
\bm X_{w_2}^a = {\rm agw}(\bm X^a), \quad
\bm X_{w}^b = {\rm agw}(\bm X^b),
\end{align}
which are used for pseudolabel generation.

Subsequently, strong augmentations are applied to the weak views to obtain
\begin{align}
\bm X_{s_1}^a = {\rm ags}(\bm X_{w_1}^a), \quad
\bm X_{s_2}^a = {\rm ags}(\bm X_{w_2}^a), \quad
\bm X_{s}^b = {\rm ags}(\bm X_{w}^b).
\end{align}

To further enhance structural diversity, we perform intra-scene and inter-scene mixing:
\begin{align}
\bm X^{aa} &= \bm{M}_1 \odot \bm X_{s_1}^{a} + (1 - \bm{M}_1) \odot \bm X^{a}_{s_2},\\
\bm X^{ab} &= \bm{M}_2 \odot \bm X_{s_2}^{a} + (1 - \bm{M}_2) \odot \bm X_{s}^{b}
\end{align}
where, \(\bm{M}_1\) and \(\bm{M}_2\) are binary masks representing two randomly sampled rectangular regions. For each rectangle, its size is first determined by randomly sampling an area ratio and an aspect ratio; subsequently, its position is uniformly and randomly placed within the image, under the constraint that the entire rectangle remains strictly inside the image boundaries. The intra-scene mixed view $\bm X^{aa}$ enriches structural variation within the same scene, while the inter-scene mixed view $\bm X^{ab}$ introduces broader contextual variability.
\subsection{Supervisions}
After feature extraction by CD-Mamba, we design an overall loss to predictions of CD-Mamba on both labeled and unlabeled data. For labeled samples, we apply the standard cross-entropy loss to enforce supervision based on ground-truth annotations. For unlabeled samples, we apply two types of weak-to-strong view-consistency losses, which constructs high-quality pseudolabels and enforces cross-view consistency to improve generalization.

For labeled data, we apply the standard supervised cross-entropy loss:
\begin{equation}
\mathcal{L}_{\text{sup}} = -\sum_{i\in\mathcal P}\sum_{j \in \mathcal{C}} t_{ij} \log y_{ij}
\end{equation}
where $ \mathcal{C} = \{0,1\}$ denotes the set of two semantic classes, $\mathcal P$ represents all pixel locations in the image,  \(t_{ij}\) is the one-hot ground truth of the \(i\)-th pixel for class \(j\), and \(y_{ij}\) is the corresponding predicted probability.

Pseudolabels of the unlabeled data are generated from the model predictions on weakly augmented inputs. The network produces a probability map \(\bm{Y} = [y_{ij}]\), and the corresponding pseudolabels are obtained as \(\hat{t}_{ij} = {\rm onehot}(y_{ij})\), where each pixel is assigned to the class with the highest predicted probability if it exceeds a confidence threshold 0.5. These pseudolabels serve as supervision for the corresponding strongly augmented samples in the unsupervised learning stage.

For the unlabeled samples, we further adopt weak-to-strong pseudo supervision, where only high-confidence pixels are used to update the model~\cite{sohn2020fixmatch}. The corresponding losses are defined by
\begin{align}
\mathcal{L}^{aa}_{\rm w2s}
&= -\sum_{i\in\mathcal P}\sum_{j \in \mathcal{C}}
\mathbb{I}(y^{aa}_{ij} > \tau) \, \hat{t}^{aa}_{ij} \log y^{aa}_{ij}\\
\mathcal{L}^{ab}_{\rm w2s}
&= -\sum_{i\in\mathcal P}\sum_{j \in \mathcal{C}}
\mathbb{I}(y^{ab}_{ij} > \tau) \, \hat{t}^{ab}_{ij} \log y^{ab}_{ij}
\end{align}
where \(\mathbb{I}(\cdot)\) denotes an indicator function selecting pixels with prediction confidence above the threshold \(\tau\). The pseudolabels passing this confidence filter are treated as hard supervision to guide the training on the mixed strong views.

At the feature level, the network produces two groups of feature representations corresponding to weakly and strongly augmented inputs, respectively:
\begin{align}
\bm{Z}^{aa}_w
&= [Z_{w,0}^{aa}; Z_{w,1}^{aa}], \quad \bm{Z}^{aa}_s = [Z_{s,0}^{aa}; Z_{s,1}^{aa}],\\
\bm{Z}^{ab}_w
&= [Z_{w,0}^{ab}; Z_{w,1}^{ab}], \quad \bm{Z}^{ab}_s = [Z_{s,0}^{ab}; Z_{s,1}^{ab}],
\end{align}
where each channel denotes the feature response of a specific semantic class. To ensure consistent semantic understanding across different augmentation strengths, a weak-to-strong view-consistency is imposed on both channels individually. Each channel is normalized by $z$-score normalization to eliminate scale differences across augmentations and stabilize learning.

We then compute the view-consistency loss by measuring the mean squared error between the weakly and strongly augmented logits after normalization, effectively encouraging their correlation\cite{xu2024structure,xia2025hierarchical}.
This loss operates at a global semantic level, aligning prediction structures across augmentation strengths and improving model robustness under limited annotations. Formally, the losses for intra-scene and inter-scene mixed samples are defined as:
\begin{align}
\mathcal{L}^{aa}_{\rm vc} &= \sum_{j \in \mathcal{C}}\| Z^{aa}_{w, j} - Z^{aa}_{s, j} \|^2, \\
\mathcal{L}^{ab}_{\rm vc} &= \sum_{j \in \mathcal{C}}\| Z^{ab}_{w, j} - Z^{ab}_{s, j} \|^2
\end{align}
where \(Z^{aa}_{w,j}\) and \(Z^{aa}_{s,j}\) denote the $z$-score normalized logits for the \(j\)-th class obtained from the weakly and strongly augmented views of the intra-scene sample, respectively; the same notation applies to the inter-scene case.

CloudMatch jointly leverages labeled and unlabeled data within a unified training framework. For labeled samples, standard cross-entropy loss is applied using ground-truth annotations. For unlabeled data, training involves two complementary objectives under both intra- and inter-scene augmentations: weak-to-strong pseudo-supervision and weak-to-strong view-consistency losses. The overall training objective is then defined as:
\begin{equation}
\mathcal{L} = \mathcal{L}_{\rm sup}
+ \lambda_{\rm w2s} (\mathcal{L}_{{\rm w2s}}^{aa}
+ \mathcal{L}_{{\rm w2s}}^{ab})
+ \lambda_{\rm vc} (\mathcal{L}_{{\rm vc}}^{aa}
+ \mathcal{L}_{{\rm vc}}^{ab})
\end{equation}
where $ \lambda_{\rm w2s}$ and $\lambda_{\rm vc} $ are hyperparameters that balance the contributions of the supervised loss, weak-to-strong pseudo-supervision loss, and weak-to-strong view-consistency loss, respectively. This joint loss formulation enables CloudMatch to effectively leverage both labeled and unlabeled data, promoting robust cloud detection even under limited annotations.
\section{Experiments}
\subsection{Experimental Setup.}
\textbf{Datasets and Evaluation Metrics.}
We conduct experiments using data collected by the Landsat-8 satellite, which was launched in 2013.
The satellite carries two core instruments: the Operational Land Imager (OLI) and the Thermal Infrared Sensor (TIRS). Landsat-8's continuous operation for over 12 years has resulted in a comprehensive multimodal dataset system, covering diverse geographic regions, seasonal variations, and cloud conditions worldwide.
These characteristics make it an ideal data source for evaluating the cross-regional generalization and complex-scenario adaptability of cloud detection algorithms.
Based on geographic representativeness, cloud diversity, and research popularity, we select three widely used remote sensing datasets for experimental analysis:
Biome~\cite{foga2017cloud}, SPARCS~\cite{USGS2016}, and RICE~\cite{lin2019remote}. The specific statistics and characteristics of these three datasets are detailed in the table~\ref{datasets}.
\begin{table}[htbp]
\centering
\caption{Statistics of the three different datasets.}\label{1}
\begin{tabular}{l|rcrr}
\toprule
Dataset   &  \# image & Resolution & \# Pixel~~~~ & \# band  \\
\midrule
Biome&
96~~~~&
$8000\times8000$&
$6.1\times10^9$&
10~~~~
\\
SPARCS&
80~~~~&
$1000\times1000$&
$0.8\times10^8$&
10~~~~
\\
RICE&
736~~~~&
$512\times512$&
$1.9\times10^8$&
3~~~~
\\
\bottomrule
\end{tabular}
\label{datasets}
\end{table}

The Biome dataset includes 96 images, each with a spatial resolution of $8000\times8000$ pixels. It evenly covers eight typical geographic environments: barren land, forest, grassland/crops, shrubland, snow/ice, urban areas, water bodies, and wetlands, with 12 images in each category. These samples span six continents and encompass diverse climate zones ranging from low-latitude equatorial to high-latitude polar regions, demonstrating significant geographical diversity and large spatial coverage. Biome is widely used to evaluate the cross-regional generalization ability of cloud detection algorithms under complex surface conditions.

The SPARCS dataset comprises 80 images, each with a spatial resolution of $1000\times1000$ pixels. Its core objective is to provide high-precision validation benchmarks for cloud and cloud-shadow masking algorithms. The dataset evenly covers typical mid- to low-latitude land surface types, including five core categories: clouds, cloud shadows, snow/ice, water bodies, and land, with approximately 16 images in each category. Scenes are distributed across global mid-low latitude regions, capturing complex scenarios with mixed thin and thick cloud cover.

The RICE dataset includes 736 image groups, each with a resolution of $512\times512$ pixels. It encompasses diverse global landscapes such as urban areas, dense vegetation, and highly reflective snow/ice regions. This dataset is particularly focused on challenging scenarios, including spectral confusion between clouds and vegetation, and between cloud shadows and urban shadows. It also systematically covers various cloud types (e.g., cirrus, stratus, cumulus) and mixtures of thin and thick clouds, which effectively tests a model's ability to discriminate cloud densities.

The Biome dataset is partitioned into 72 geographic scenes for training and 24 scenes for testing, ensuring spatial separation between the training and test sets to prevent data leakage. The input consists of the standard red, green, and blue (RGB) spectral bands to maintain compatibility across imagery from different satellite sensors. The raw pixel values are first linearly mapped to the range \([0, 255]\) and then normalized using a standardization procedure widely adopted in the remote sensing community to enhance model generalization. To balance computational efficiency with sufficient contextual information, each image is divided into non-overlapping patches of size \(384 \times 384\). This yields a total of 10,368 training samples and 7,682 test samples from the Biome dataset.

Under the semi-supervised learning setting, we employ a hierarchical sampling strategy to construct labeled subsets at different annotation ratios (i.e., 1/4, 1/8, and 1/16). Specifically, we first randomly select 1/4 of the full training set (10,368 samples) as the labeled subset for the 1/4 ratio. From this subset, we randomly sample 50\% (equivalent to 1/8 of the full training set) to form the labeled set for the 1/8 ratio. Similarly, the 1/16 labeled set is obtained by further halving the 1/8 subset. In each configuration, the remaining training samples are treated as unlabeled data for semi-supervised learning. This recursive sampling scheme ensures both spatial representativeness and experimental reproducibility across different labeling budgets. Importantly, the test set remains identical across all labeling ratios, guaranteeing fair and comparable evaluation across experimental settings. This design enables a systematic assessment of the model’s performance under limited annotation scenarios and its sensitivity to labeling efficiency.

Since the SPARCS and RICE datasets contain a limited number of samples, we use them entirely as test sets, without any train/validation split, to avoid validation bias and specifically evaluate the model’s cross-dataset generalization capability. Both SPARCS and RICE are kept at their original resolutions without cropping and are evaluated on full images to assess the model’s generalization across varying spatial scales.

The performance of the proposed method was evaluated using mean Intersection over Union (mIoU) and accuracy (ACC), which quantify segmentation overlap and pixel-wise correctness, respectively. The calculating formulas are as follows:
\begin{align}
\mathrm{IoU}_0
&=\frac{\mathrm{TN}}{\mathrm{TN}+\mathrm{FN}+\mathrm{FP}}\notag\\
\mathrm{IoU}_1
&=\frac{\mathrm{TP}}{\mathrm{TP}+\mathrm{FP}+\mathrm{FN}}\notag\\
\mathrm{mIoU}
&=\frac{\mathrm{IoU}_0+\mathrm{IoU}_1}{2}, \\
\mathrm{ACC}
&=\frac{\mathrm{TP}+\mathrm{TN}}{\mathrm{TP}+\mathrm{TN}+\mathrm{FP}+\mathrm{FN}} \,,
\end{align}
where true positives (TP) denote correctly classified cloud pixels; true negatives (TN), correctly classified non-cloud pixels; false positives (FP), non-cloud pixels misclassified as cloud; and false negatives (FN), cloud pixels misclassified as non-cloud. The metrics are evaluated over all pixels in the test set.

\textbf{Backbone Selection.} To evaluate the efficacy of different architectures in addressing the core challenges of cloud detection, namely capturing multi-scale structures and fine boundaries, we compared CD-Mamba~\cite{xueJARS2025} with DeepLab~v3+~\cite{Chen_2018_ECCV}, a widely adopted backbone in semantic segmentation. For a fair comparison that isolates the architectural benefits, both models were trained from scratch without pre-trained weights.

\begin{figure}[!t]
\centering
\includegraphics[width=0.9\linewidth]{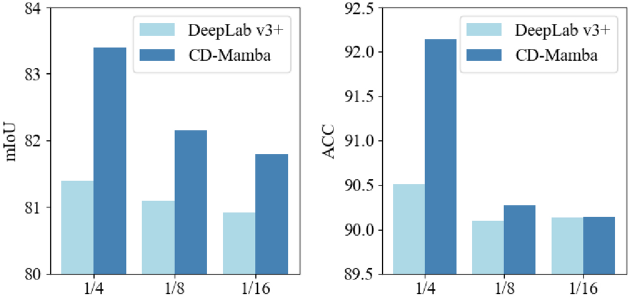}
\caption{Comparative experimental results of different backbones on the Biome dataset.}
\label{BACKBONES}
\end{figure}

Figure~\ref{BACKBONES} shows that CD-Mamba consistently outperforms DeepLab~v3+ across all labeled data ratios. Its advantage is particularly evident in capturing multi-scale cloud regions and delineating complex boundaries. By effectively modeling long-range dependencies, CD-Mamba integrates contextual information over large areas, while its dynamic routing mechanism enhances sensitivity to thin cloud edges, thereby addressing the dual challenges of scale variation and edge clarity. These results confirm CD-Mamba's superior feature extraction capability for remote sensing cloud detection, and we therefore adopt it as the backbone network for CloudMatch.

Moreover, to comprehensively evaluate computational efficiency, we further compare the number of parameters, floating-point operations (FLOPs), and actual inference time of the models under identical experimental settings , as shown in Table \ref{modelparam}. CD-Mamba has a significantly smaller model size compared to other backbones, yet its inference time does not suffer a substantial increase, demonstrating superior efficiency and practicality. These results confirm that CD-Mamba is better suited for capturing long-range dependencies, and thus we adopt it as the backbone network for CloudMatch.
\begin{table*}[!t]
\centering
\caption{FLOPs, parameter count, and average inference time comparison of detection models.}
\label{modelparam}  
\begin{tabular}{p{2.5cm}rrrr}  
\toprule
Models & FLOPS (GFLOPs) & Param (MB) & Inference Time (ms) \\
\midrule
UNet            & 90.428~~~~~~~~~~~           & 17.263~~~~~~          & 9.393~~~~~~~~~~~        \\
DeepLab v2      & 104.249~~~~~~~~~~~            & 42.574~~~~~~          & 12.491~~~~~~~~~~~
\\
DeepLab v3+     & 106.993~~~~~~~~~~~            & 40.471~~~~~~         & 11.982~~~~~~~~~~~             \\
CD-Mamba         & 2.020~~~~~~~~~~~   & 0.050~~~~~~ & 16.573~~~~~~~~~~~           \\
\bottomrule
\end{tabular}
\end{table*}

\textbf{Implementation Details.} We conducted all experiments on a system running Ubuntu 20.04.6, using Python 3.10 to develop the models. The training was performed on an NVIDIA RTX 3090 GPU with a batch size of 4 for 80 epochs.

For the semi-supervised learning baselines, we adhered strictly to their original experimental setups, including the optimization algorithms, data preprocessing procedures, and hyperparameter values, to guarantee fair comparisons.

For data augmentation, we adopt a dual-branch augmentation strategy consistent with baselines~\cite{Yang_2023_CVPR,sun2024corrmatch,tain2024}. The random scaling factor is sampled from the range \([0.5, 2.0]\), horizontal flipping is applied with a probability of 0.5, color jittering with an intensity of 0.5 and an application probability of 0.8, grayscale conversion with a probability of 0.2, and Gaussian blur with a probability of 0.5, where the standard deviation is uniformly sampled from the interval \([0.1, 2.0]\). In our approach, the mixing operation is applied with probabilities of 0.5 and 0.8 in cross-scene mixing and within-scene mixing, respectively. During the mixing process, the area ratio of the cropped region is uniformly sampled from the interval \([0.02, 0.4]\), and the aspect ratio is randomly sampled from the range \([0.3, 1/0.3]\).

In the proposed method, the loss weighting coefficients were empirically determined through preliminary experiments and set as follows: $\lambda_{\rm w2s}=0.5$, and $\lambda_{\rm vc}=0.5$ to balance the contribution of each component. 

Furthermore, we employ an adaptive confidence thresholding strategy~\cite{wang2023freematch} to select pixels whose predicted confidence scores exceed a dynamic threshold $\tau$. This strategy effectively suppresses the adverse impact of low-quality pseudolabels, thereby improving detection performance in semi-supervised or weakly-supervised settings.
\subsection{Experimental Results.}
\textbf{Quantitative Evaluation.} We compared the performance of CloudMatch with state-of-the-art semi-supervised segmentation methods, including CPS~\cite{DBLP:journals/corr/abs-2106-01226}, DSSN~\cite{tain2024}, UniMatch~\cite{Yang_2023_CVPR}, CorrMatch~\cite{sun2024corrmatch}, and a semi-supervised method specifically applied to cloud detection, SSCDnet~\cite{sscd}. To ensure fairness and consistency in evaluation, all methods were trained on Biome using identical training strategies and parameter settings. During training, we consistently used CD-Mamba as the network backbone for CPS, DSSN, UniMatch, CorrMatch, and our approach CloudMatch.

The experimental results on Biome are shown in Table~\ref{Biometest1}. To ensure a fair comparison of learning strategies, all methods in this experiment are implemented using CD-Mamba as the shared backbone network. Under this unified backbone setting, CloudMatch consistently achieves the best performance across all evaluation metrics and label splits. Specifically, for the 1/4, 1/8, and 1/16 labeled data settings, CloudMatch outperforms the second-best method by +2.03\%, +2.75\%, and +3.11\% in mIoU, and +1.47\%, +0.71\%, and +0.88\% in in ACC, respectively. In addition to detection performance, Table~\ref{Biometest1} also reports the memory consumption and per-epoch training time.

To further demonstrate CloudMatch's superiority, we compare CloudMatch with fully supervised network models trained only on labeled data, achieving mIoU (83.69) and ACC (92.60). Under the 1/4 split, CloudMatch’s mIoU and ACC differed by only 0.3\% and 0.46\%, respectively, from these fully supervised results.

To evaluate the effectiveness of CloudMatch under realistic and fair comparison settings, we retain the default backbone architectures of all competing semi-supervised segmentation methods and retrain them on the Biome dataset using identical training protocols and hyperparameter settings. The quantitative results are reported in Table~\ref{Biometest2}. As shown in Table~\ref{Biometest2}, CloudMatch consistently achieves superior performance across all label ratios, despite different methods adopting different backbone architectures. These results indicate that the performance gains of CloudMatch are primarily attributed to the proposed learning strategy rather than reliance on a specific backbone.

These results highlight CloudMatch’s high stability and consistency under limited annotations, accurately segmenting cloud regions across diverse scenarios, which is crucial for practical remote sensing applications.
\begin{table*}[!t]
\centering
\caption{Cloud detection performance on Biome with CD-Mamba as a backbone for all methods.}\label{Biometest1}
\begin{tabular}{p{2cm}|p{2.63cm}p{2.63cm}p{2.63cm}p{2.63cm}}
\toprule
Method  &  {1/4 (2592)} & {1/8 (1296)} & {1/16 (648)} & GPU ~~~~~~time \\
\end{tabular}
\begin{tabular}{p{2cm}|p{1.1cm}p{1.1cm}p{1.1cm}p{1.1cm}p{1.1cm}p{1.1cm}|p{1.1cm}p{1.1cm}}
&mIoU&ACC&mIoU&ACC&mIoU& ACC& (MB)&(min)\\
\midrule
CPS&76.27 & 86.94&73.16&86.84&72.64&86.75&6424&36\\
DSSN&79.44 &89.06 &79.39&89.51& 76.97 &87.40&10068&18\\
UniMatch&\textbf{\textit{81.36}}&\textbf{\textit{90.67}}&78.58&\textbf{\textit{89.56}}&\textbf{\textit{78.69}}&\textbf{\textit{89.26}}&11238&16\\
CorrMatch& 80.41&88.86&\textbf{\textit{79.40}}&88.74&78.25&88.84&8988&15\\
\midrule
CloudMatch&\textbf{83.39}&\textbf{92.14}&\textbf{82.15}&\textbf{90.27}&\textbf{81.80}&\textbf{90.14}&17892&23\\
\end{tabular}
\begin{tabular}{p{8.2cm}p{1.1cm}p{1.1cm}|p{1.1cm}p{1.1cm}}
\midrule
Train using all labeled images &83.69 & 92.60&18044 &31\\
\bottomrule
\end{tabular}
\end{table*}

\begin{table*}[!t]
\centering
\caption{Cloud detection performance on the Biome using each method’s default backbone.}\label{Biometest2}
\begin{tabular}{p{2cm}|p{2.63cm}p{2.63cm}p{2.63cm}p{2.63cm}}
\toprule
Method  &  {1/4 (2592)} & {1/8 (1296)} & {1/16 (648)} & GPU ~~~~~~time \\
\end{tabular}

\begin{tabular}{p{2cm}|p{1.1cm}p{1.1cm}p{1.1cm}p{1.1cm}p{1.1cm}p{1.1cm}|p{1.1cm}p{1.1cm}}
&mIoU&ACC&mIoU&ACC&mIoU& ACC& (MB)&(min)\\
\midrule
SSCDnet&75.76  & 87.31&74.21 & 86.99 & 72.17 & 86.71 &11500&14\\
DSSN&\textbf{\textit{81.26}}&\textbf{\textit{90.42}}&78.89&88.85&77.08&87.87&16372& 17\\
UniMatch&79.239&90.14&\textbf{\textit{80.042}}&\textbf{\textit{90.03}}&\textbf{\textit{80.24}}&\textbf{\textit{89.60}}&12540&12\\
CorrMatch&76.70 &87.66&77.65&86.54&76.56&86.03&18016&24\\
\midrule
CloudMatch&\textbf{83.39}&\textbf{92.14}&\textbf{82.15}&\textbf{90.27}&\textbf{81.80}&\textbf{90.14}&17892&23\\
\bottomrule
\end{tabular}
\end{table*}
Although all three datasets are derived from Landsat 8, the generated images exhibit significant visual discrepancies due to different processing levels and distinctive color mapping strategies. Therefore, for cross-dataset cross-validation, we normalized the color space of the training sets by unifying color mapping strategies. Additionally, significant discrepancies exist in the image mask annotations made by different researchers across datasets, which provide important research value for cross-dataset inductive experiments.

\begin{figure}[!t]
    \centering
    \includegraphics[width=0.7\linewidth]{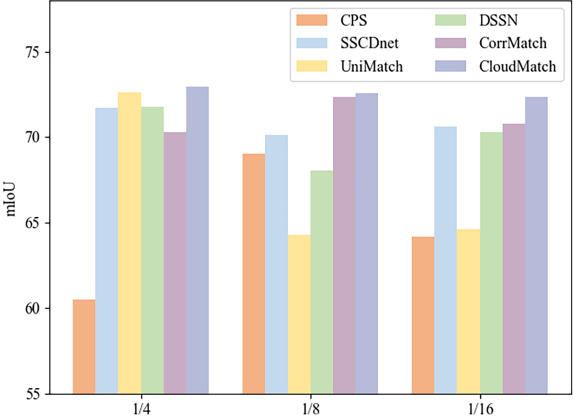}
    \caption{Comparative experimental results on the RICE dataset.}
    \label{Rice}
\end{figure}

\begin{figure}[!t]
    \centering
    \includegraphics[width=0.7\linewidth]{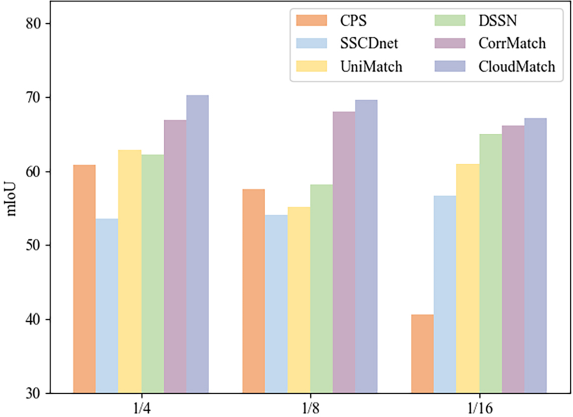}
    \caption{Comparative experimental results on the SPARCS dataset.}
    \label{SPARS}
\end{figure}

To comprehensively verify the generalization performance of the CloudMatch, we test the trained model on the SPARCS dataset and RICE dataset, respectively. The experimental results are detailed in Figures~\ref{Rice} and \ref{SPARS}, which show that the CloudMatch has superior generalization performance on both datasets.


\begin{figure}[!t]
\begin{center}
\includegraphics[width=0.48\textwidth]{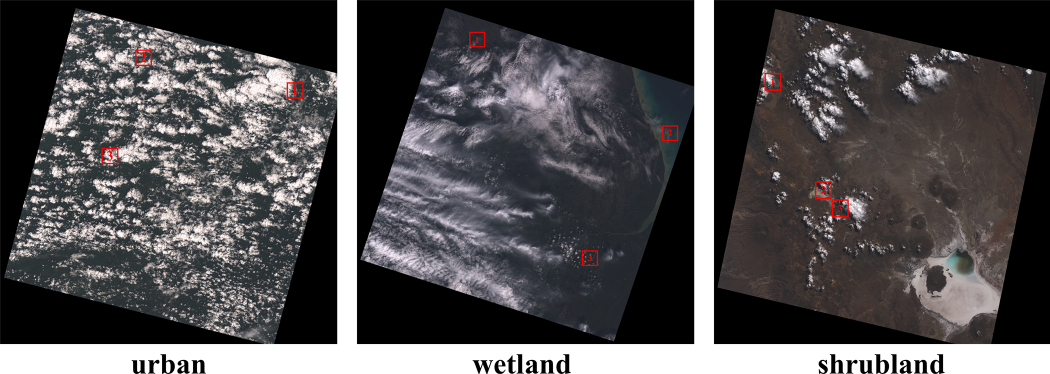}
\caption{Big picture of the Biome dataset, including: urban, wetland, shrubland.}
\label{Biome}
\end{center}
\end{figure}

\begin{figure*}[!t]
\begin{center}
\includegraphics[width=0.98\textwidth]{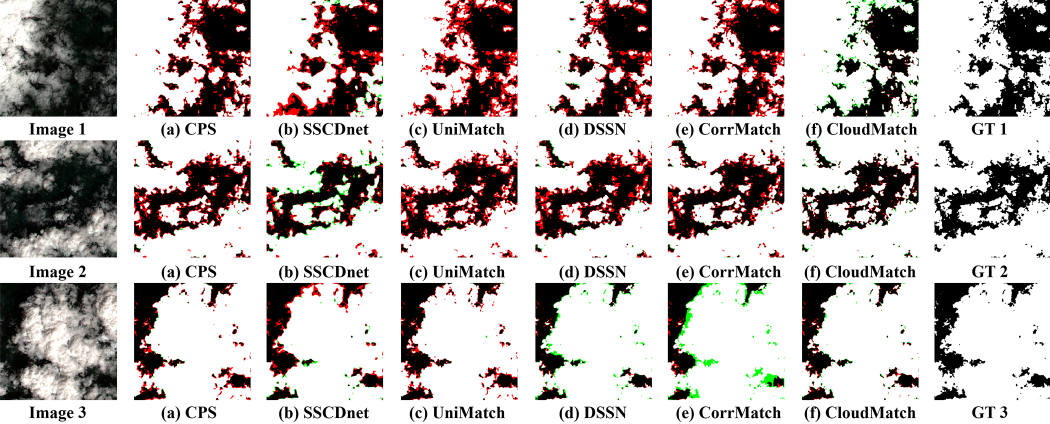}
\caption{In the urban scene of Biome, there is a snow-free area with moderate cloud cover.}
\label{Biome1}
\end{center}
\end{figure*}

\begin{figure*}[!t]
\begin{center}
\includegraphics[width=0.98\textwidth]{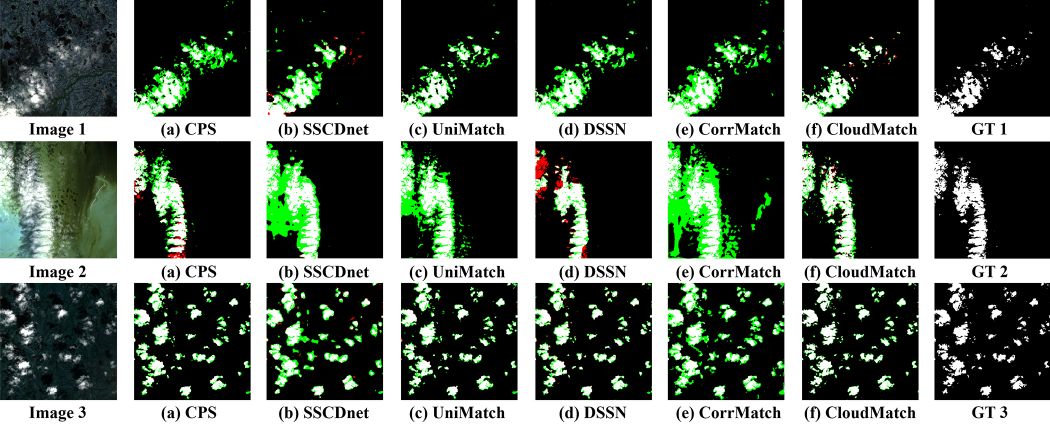}
\caption{In the wetland scene of Biome, there is a snow-free area with moderate cloud cover.}
\label{Biome2}
\end{center}
\end{figure*}

\begin{figure*}[!t]
\begin{center}
\includegraphics[width=0.98\textwidth]{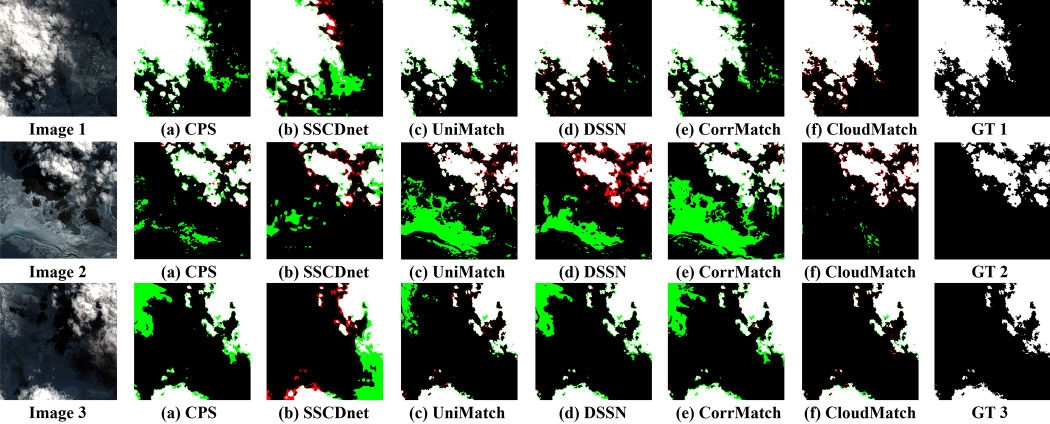}
\caption{In the shrubland scene of Biome, there is a snow-covered area with low cloud cover.}
\label{Biome3}
\end{center}
\end{figure*}

\textbf{Qualitative Evaluation.} Figure~\ref{Biome} shows three large-scale images randomly sampled from the Biome dataset, covering urban, wetland and shrubland scenes with varying cloud amounts and geographical conditions. Figures~\ref{Biome1},~\ref{Biome2}, and~\ref{Biome3} provide qualitative comparisons between CloudMatch and other methods in three representative Biome scenes selected from these images. In the visual results, red markers indicate missed detections (i.e., undetected cloud areas), while green markers represent false positives (i.e., non-cloud regions misidentified as clouds).

To verify the effectiveness of CloudMatch in detecting clouds across diverse scenes, we select representative regions from three distinct environments: (1) snow-free urban areas with medium cloud coverage, (2) snow-free wetland areas with medium cloud coverage, and (3) snowy shrubland areas with low cloud coverage. Figure~\ref{Biome1} presents the detection performance of CloudMatch in the urban scene. In this setting, clouds are relatively simple, mainly consisting of thick and thin clouds. As shown, CloudMatch achieves high detection accuracy, particularly in identifying both thick clouds and challenging thin clouds and cloud boundaries. Compared to other methods, CloudMatch preserves fine-grained details and produces more precise boundary delineation.

Figure~\ref{Biome2} illustrates the detection results in the wetland scene. Compared to the urban environment, wetlands often contain rain-affected regions and bright surfaces that resemble cloud structures.
In Images 1 and 2, such areas pose challenges for other methods, leading to a higher rate of false positives. In contrast, CloudMatch effectively distinguishes true clouds from cloud-like features, significantly reducing misclassifications and exhibiting strong robustness in boundary handling.

Figure~\ref{Biome3} displays the detection results in the shrubland scene, which includes mixed ice-water regions and highly reflective surfaces that increase detection complexity. As observed from the figure, SSCDnet, specifically designed for cloud detection, performs better than general semi-supervised models, achieving relatively lower error rates. However, CloudMatch delivers the best overall performance, accurately differentiating reflective ice/snow regions from actual clouds, while also producing clearer and more detailed cloud boundaries.

These results demonstrate that CloudMatch can effectively identify and segment cloud regions under varied scenarios, enhancing reliability for real-world remote sensing applications.

CloudMatch's powerful detection performance is attributed to its synergistic modules: (1) the view consistency loss module enhances cross-scene robustness through weak-strong view alignment, (2) the inter- and intra-scene mixing augmentation increases sample feature diversity through intra-scene structural variation and inter-scene contextual mixing, and (3) the CD-Mamba architecture captures global cloud distribution and fine-grained textures with its sequential modeling capability. The synergistic effect of these technologies allows the model to maintain high detection accuracy in complex single images and demonstrate robust performance in challenging scenarios such as rain-snow coexistence, high-brightness regions, and visually similar areas.

\begin{figure*}[!t]
\begin{center}
\includegraphics[width=1\textwidth]{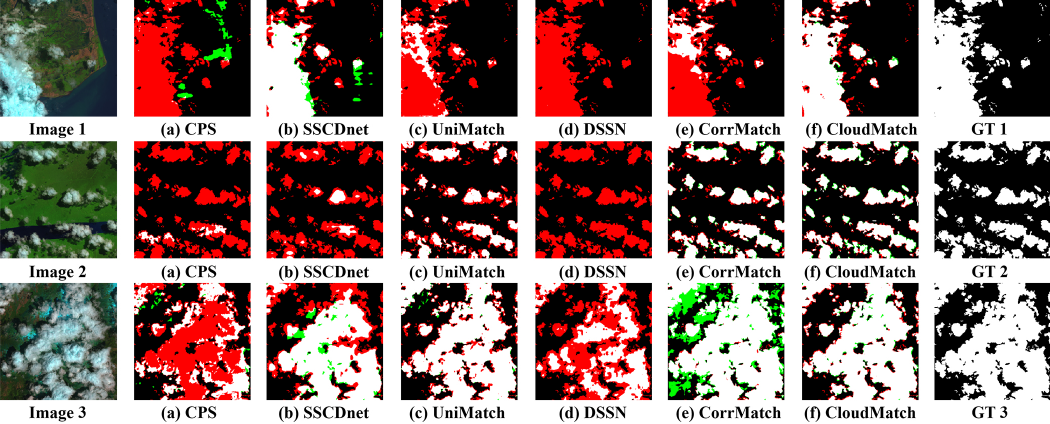}
\caption{Comparative results of different detection methods on three randomly selected images from the SPARCS dataset.}
\label{spars}
\end{center}
\end{figure*}

\begin{figure*}[!t]
\begin{center}
\includegraphics[width=1\textwidth]{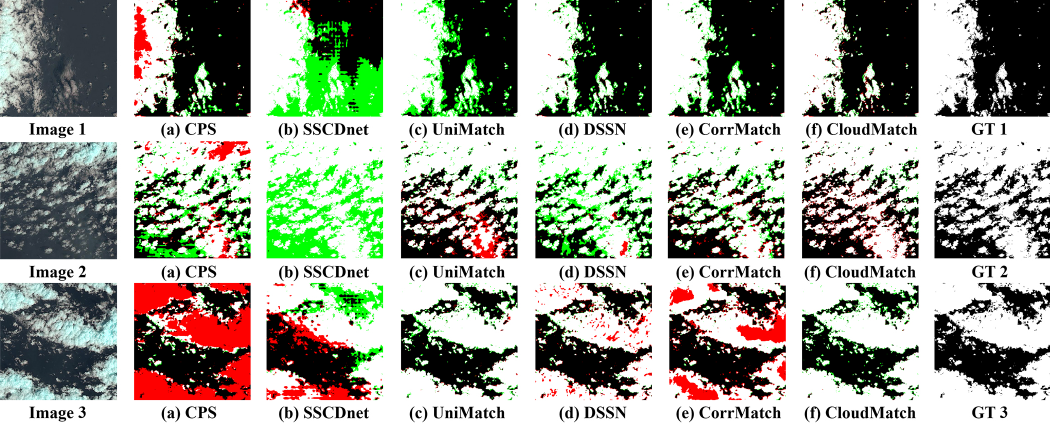}
\caption{Comparative results of different detection methods on three randomly selected images from the RICE dataset.}
\label{rice}
\end{center}
\end{figure*}

To further validate the cross-dataset generalization ability of CloudMatch, we conduct qualitative experiments on the SPARCS and RICE datasets, as shown in Figures~\ref{spars} and~\ref{rice}, and carefully selected three typical scenarios for comparison: First, we choose edge-region images with complex cloud boundary information (first row of Figures~\ref{spars} and~\ref{rice}). In such scenarios that demand high algorithm processing capabilities, CloudMatch achieves superior boundary segmentation accuracy, precisely delineating intricate cloud edges.
Second, we use images with concentrated cloud blocks and moderate cloud cover (second row of Figures~\ref{spars} and~\ref{rice}). In these scenes, CloudMatch not only achieves the lowest false positive rate compared to other methods but also precisely handles the connection areas between clouds, greatly improving the integrity and accuracy of detection. Finally, we select images with abundant cloud cover, large cloud blocks, and complex features such as highlighted areas (third row of Figures~\ref{spars} and~\ref{rice}). Even in these highly challenging scenarios, CloudMatch maintains optimal performance, effectively suppressing interference and completely capturing cloud morphology. These experiments demonstrate that CloudMatch maintains low false positive and false negative rates across various cloud densities, boundary complexities, and ground interference conditions, showcasing strong robustness and detection performance.

\textbf{Ablation Study.}
We conduct extensive ablation studies to systematically evaluate the effectiveness of each core module. Experiments are performed under the 1/4 labeled data setting, using mIoU and ACC as evaluation metrics. The results are presented in Table \ref{Ablation}.

\begin{table}[!t]
\centering
\caption{Ablation study of loss functions under 1/4 data partition.}
\begin{tabular}{l|cc}
\toprule
Setting & mIoU  & ACC\\
\midrule
Cloudmatch w/o $\mathcal{L}_{{\rm vc}}$ &82.27 &90.56\\
Cloudmatch w/o Inter-Scene Mix &82.58& 90.86\\
Cloudmatch w/o Intra-Scene Mix &80.74& 89.46\\
\hline
Cloudmatch&\textbf{83.39} &\textbf{92.14} \\
\bottomrule
\end{tabular}
\label{Ablation}
\end{table}
The full CloudMatch model, which integrates all proposed modules, achieved the highest performance, with an mIoU of 83.39\% and ACC of 92.14\%. When the view-consistency loss module was removed, mIoU dropped by 1.12\% and ACC decreased by 1.58\%. The view consistency loss enhances model robustness in complex conditions such as rain, fog, and high brightness by aligning class-level features between weakly and strongly augmented views, thereby reducing misclassification caused by spectral confusion. Inter-scene mixing enriches the diversity of cloud patterns by blending structural and textural features from different scenes, enabling the model to recognize rare or unseen cloud types. Intra-scene mixing further helps the model adapt to domain shifts arising from geographical or imaging condition variations, maintaining stable performance even in regions outside the training distribution. Together, these mechanisms improve the model's generalization ability across complex scenes, diverse cloud patterns, and varying domain conditions. Consistently, removing Inter-Scene Mix led to a drop of 0.81\% in mIoU and 1.28\% in ACC, while removing Intra-Scene Mix caused an even larger decline, with both mIoU and ACC decreasing by more than 2.6\%.

These results validate that each module plays a critical and complementary role,
and their integration enables CloudMatch to maintain robust and accurate cloud detection under limited annotations.
\section{Conclusion}
In this paper, we present CloudMatch, a unified semi-supervised framework for remote sensing cloud detection. Built upon view-consistency learning, CloudMatch leverages unlabeled data through two key components: (1) a weak-to-strong view-consistency loss that enforces class-level semantic alignment between weakly and strongly augmented views, enhancing feature robustness; and (2) a dual scene-mixing augmentation module combining inter-scene patch mixing with intra-scene spatial transformations to better capture the complex appearance of real-world clouds.
To model long-range dependencies in cloud structures, we integrate the CD-Mamba, enabling more accurate discrimination of clouds from confusable surfaces such as snow or water bodies. CloudMatch is an effective method for accurate and annotation-efficient cloud detection in remote sensing.

While CloudMatch demonstrates strong performance under limited supervision, it has several limitations. Like most pseudolabel-based semi-supervised methods, CloudMatch remains dependent on the quality of predictions, and performance may degrade under extremely low annotation ratios or severe domain shifts. The use of multiple augmented views and mixing operations increases training-time computational overhead, although inference remains unchanged. The current design is primarily evaluated on binary cloud detection, and extending the framework to multi-class or fine-grained cloud categorization requires further investigation. These limitations point to several promising directions for future work, including adaptive scene partitioning, confidence-aware pseudo-label refinement, and extension to multi-class cloud understanding tasks.
\section*{Disclosures}
The authors declare that there are no financial interests, commercial affiliations, or other potential conflicts of interest that could have influenced the objectivity of this research or the writing of this paper.
\section*{Code, Data, and Materials Availability}
The source code is available at \url{https://github.com/kunzhan/CloudMatch}. The download links for the two datasets are provided in README.md of the GitHub repository.

%% file: cloudMatch.bbl
\begin{thebibliography}{38}
\providecommand{\natexlab}[1]{#1}
\providecommand{\url}[1]{\texttt{#1}}
\expandafter\ifx\csname urlstyle\endcsname\relax
  \providecommand{\doi}[1]{doi: #1}\else
  \providecommand{\doi}{doi: \begingroup \urlstyle{rm}\Url}\fi

\bibitem[Abuduweili et~al.(2021)Abuduweili, Li, Shi, Xu, and
  Dou]{Abuduweili_2021_CVPR}
Abulikemu Abuduweili, Xingjian Li, Humphrey Shi, Cheng-Zhong Xu, and Dejing
  Dou.
\newblock Adaptive consistency regularization for semi-supervised transfer
  learning.
\newblock In \emph{CVPR}, pages 6923--6932, 2021.

\bibitem[Berthelot et~al.(2019)Berthelot, Carlini, Goodfellow, Papernot,
  Oliver, and Raffel]{NEURIPS2019_1cd138d0}
David Berthelot, Nicholas Carlini, Ian Goodfellow, Nicolas Papernot, Avital
  Oliver, and Colin~A Raffel.
\newblock {MixMatch}: A holistic approach to semi-supervised learning.
\newblock In \emph{NeurIPS}, 2019.

\bibitem[Chen et~al.(2023)Chen, Tao, Fan, Wang, Wang, Schiele, Xie, Raj, and
  Savvides]{softmatch}
Hao Chen, Ran Tao, Yue Fan, Yidong Wang, Jindong Wang, Bernt Schiele, Xing Xie,
  Bhiksha Raj, and Marios Savvides.
\newblock {SoftMatch}: Addressing the quantity-quality trade-off in
  semi-supervised learning.
\newblock In \emph{ICLR}, 2023.

\bibitem[Chen et~al.(2018)Chen, Zhu, Papandreou, Schroff, and
  Adam]{Chen_2018_ECCV}
Liang-Chieh Chen, Yukun Zhu, George Papandreou, Florian Schroff, and Hartwig
  Adam.
\newblock {Encoder-Decoder} with atrous separable convolution for semantic
  image segmentation.
\newblock In \emph{ECCV}, pages 801--818, 2018.

\bibitem[Chen et~al.(2021)Chen, Yuan, Zeng, and
  Wang]{DBLP:journals/corr/abs-2106-01226}
Xiaokang Chen, Yuhui Yuan, Gang Zeng, and Jingdong Wang.
\newblock Semi-supervised semantic segmentation with cross pseudo supervision.
\newblock In \emph{CVPR}, pages 2613--2622, 2021.

\bibitem[Foga et~al.(2017)Foga, Scaramuzza, Guo, Zhu, Dilley, Beckmann,
  Schmidt, Dwyer, {Joseph Hughes}, and Laue]{foga2017cloud}
Steve Foga, Pat~L. Scaramuzza, Song Guo, Zhe Zhu, Ronald~D. Dilley, Tim
  Beckmann, Gail~L. Schmidt, John~L. Dwyer, M. {Joseph Hughes}, and Brady Laue.
\newblock Cloud detection algorithm comparison and validation for operational
  {Landsat} data products.
\newblock \emph{Remote Sensing of Environment}, 194:\penalty0 379--390, 2017.

\bibitem[Guo et~al.(2022{\natexlab{a}})Guo, Xu, Zeng, Liu, and Zhu]{sscd}
Jianhua Guo, Qingsong Xu, Yue Zeng, Zhiheng Liu, and Xiaoxiang Zhu.
\newblock Semi-supervised cloud detection in satellite images by considering
  the domain shift problem.
\newblock \emph{Remote Sensing}, 14\penalty0 (11), 2022{\natexlab{a}}.

\bibitem[Guo et~al.(2022{\natexlab{b}})Guo, Yang, Yue, Liu, and Li]{9570362}
Jianhua Guo, Jingyu Yang, Huanjing Yue, Xin Liu, and Kun Li.
\newblock Unsupervised domain-invariant feature learning for cloud detection of
  remote sensing images.
\newblock \emph{IEEE Transactions on Geoscience and Remote Sensing},
  60:\penalty0 3120001, 2022{\natexlab{b}}.

\bibitem[He and Hong(2025)]{He_2025_ICCV}
Hongyang He and Yundi Hong.
\newblock {TrustMatch}: Mitigating pseudo-label bias in semi-supervised
  learning with trust-aware refinement.
\newblock In \emph{ICCV Workshop}, pages 594--603, 2025.

\bibitem[He et~al.(2022)He, Sun, Yan, and Fu]{dabnet}
Qibin He, Xian Sun, Zhiyuan Yan, and Kun Fu.
\newblock {DABNet}: Deformable contextual and boundary-weighted network for
  cloud detection in remote sensing images.
\newblock \emph{IEEE Transactions on Geoscience and Remote Sensing},
  60:\penalty0 3045474, 2022.

\bibitem[Hughes and Hayes(2014)]{USGS2016}
M.~Joseph Hughes and Daniel~J. Hayes.
\newblock Automated detection of cloud and cloud shadow in single-date
  {Landsat} imagery using neural networks and spatial post-processing.
\newblock \emph{Remote Sensing}, 6\penalty0 (6):\penalty0 4907--4926, 2014.

\bibitem[Kang~Wu and Ren(2023)]{doi:10.1080/15481603.2022.2147298}
Xinrong~Lyu Kang~Wu, Zunxiao~Xu and Peng Ren.
\newblock Cross-supervised learning for cloud detection.
\newblock \emph{GIScience \& Remote Sensing}, 60\penalty0 (1):\penalty0
  2147298, 2023.

\bibitem[Li et~al.(2024)Li, Xue, Zhao, Ge, Min, Su, and Zhan]{hrcloudnet}
Jingsheng Li, Tianxiang Xue, Jiayi Zhao, Jingmin Ge, Yufang Min, Wei Su, and
  Kun Zhan.
\newblock {High-resolution cloud detection network}.
\newblock \emph{Journal of Electronic Imaging}, 33\penalty0 (4):\penalty0
  043027, 2024.

\bibitem[Li et~al.(2016)Li, Zhang, Shao, Li, Hong, Liu, Li, Wei, Li, Li,
  et~al.]{li2016remote}
Zhengqiang Li, Ying Zhang, Jie Shao, Baosheng Li, Jin Hong, Dong Liu, Donghui
  Li, Peng Wei, Wei Li, Lei Li, et~al.
\newblock Remote sensing of atmospheric particulate mass of dry {PM2.5} near
  the ground: Method validation using ground-based measurements.
\newblock \emph{Remote Sensing of Environment}, 173:\penalty0 59--68, 2016.

\bibitem[Li et~al.(2023)Li, Pan, Zhang, Wang, and Liu]{rs15082040}
Zongrui Li, Jun Pan, Zhuoer Zhang, Mi Wang, and Likun Liu.
\newblock {MTCSNet}: Mean teachers cross-supervision network for
  semi-supervised cloud detection.
\newblock \emph{Remote Sensing}, 15\penalty0 (8), 2023.

\bibitem[Lin et~al.(2019)Lin, Xu, Wang, Wang, Sun, and Fu]{lin2019remote}
Daoyu Lin, Guangluan Xu, Xiaoke Wang, Yang Wang, Xian Sun, and Kun Fu.
\newblock A remote sensing image dataset for cloud removal.
\newblock \emph{arXiv:1901.00600}, 2019.

\bibitem[Ling et~al.(2021)Ling, Zhang, and Lin]{rs13224708}
Jing Ling, Hongsheng Zhang, and Yinyi Lin.
\newblock Improving urban land cover classification in cloud-prone areas with
  polarimetric sar images.
\newblock \emph{Remote Sensing}, 13\penalty0 (22), 2021.

\bibitem[Liu et~al.(2024)Liu, Luo, Huang, Wu, Jiang, and Zhang]{10769516}
Ruizhong Liu, Tingzhang Luo, Shaoguang Huang, Yuwei Wu, Zhen Jiang, and Hongyan
  Zhang.
\newblock {CrossMatch}: Cross-view matching for semi-supervised remote sensing
  image segmentation.
\newblock \emph{IEEE Transactions on Geoscience and Remote Sensing},
  62:\penalty0 1--15, 2024.

\bibitem[Lu et~al.(2025)Lu, Li, Jiao, Liu, Liu, Ma, and Yang]{10891590}
Xiaoqiang Lu, Lingling Li, Licheng Jiao, Xu Liu, Fang Liu, Wenping Ma, and
  Shuyuan Yang.
\newblock Uncertainty-aware semi-supervised learning segmentation for remote
  sensing images.
\newblock \emph{IEEE Transactions on Multimedia}, 27:\penalty0 5548--5562,
  2025.

\bibitem[Mai et~al.(2024)Mai, Sun, Zhang, and Wu]{Mai_2024_CVPR}
Huayu Mai, Rui Sun, Tianzhu Zhang, and Feng Wu.
\newblock {RankMatch}: Exploring the better consistency regularization for
  semi-supervised semantic segmentation.
\newblock In \emph{CVPR}, pages 3391--3401, 2024.

\bibitem[Olsson et~al.(2021)Olsson, Tranheden, Pinto, and
  Svensson]{DBLP:classmix}
Viktor Olsson, Wilhelm Tranheden, Juliano Pinto, and Lennart Svensson.
\newblock {ClassMix}: Segmentation-based data augmentation for semi-supervised
  learning.
\newblock In \emph{WACV}, pages 1369--1378, 2021.

\bibitem[Sohn et~al.(2020)Sohn, Berthelot, Carlini, Zhang, Zhang, Raffel,
  Cubuk, Kurakin, and Li]{sohn2020fixmatch}
Kihyuk Sohn, David Berthelot, Nicholas Carlini, Zizhao Zhang, Han Zhang,
  Colin~A Raffel, Ekin~Dogus Cubuk, Alexey Kurakin, and Chun-Liang Li.
\newblock {FixMatch}: Simplifying semi-supervised learning with consistency and
  confidence.
\newblock In \emph{NeurIPS}, pages 596--608, 2020.

\bibitem[Sun et~al.(2024)Sun, Yang, Zhang, Cheng, and Hou]{sun2024corrmatch}
Boyuan Sun, Yuqi Yang, Le Zhang, Ming-Ming Cheng, and Qibin Hou.
\newblock {CorrMatch}: Label propagation via correlation matching for
  semi-supervised semantic segmentation.
\newblock In \emph{CVPR}, pages 3097--3107, 2024.

\bibitem[Tian et~al.(2023)Tian, Zhang, Zhang, and Zhan]{tain2024}
Zhibo Tian, Xiaolin Zhang, Peng Zhang, and Kun Zhan.
\newblock Improving semi-supervised semantic segmentation with dual-evel
  siamese structure network.
\newblock In \emph{ACM Multimedia}, page 4200–4208, 2023.

\bibitem[Wang et~al.(2025)Wang, Sun, Chen, Hong, and Han]{11062866}
Shanwen Wang, Xin Sun, Changrui Chen, Danfeng Hong, and Jungong Han.
\newblock Semi-supervised semantic segmentation for remote sensing images via
  multiscale uncertainty consistency and cross-teacher–student attention.
\newblock \emph{IEEE Transactions on Geoscience and Remote Sensing},
  63:\penalty0 1--15, 2025.

\bibitem[Wang et~al.(2023)Wang, Chen, Heng, Hou, Fan, Wu, Wang, Savvides,
  Shinozaki, Raj, et~al.]{wang2023freematch}
Yidong Wang, Hao Chen, Qiang Heng, Wenxin Hou, Yue Fan, Zhen Wu, Jindong Wang,
  Marios Savvides, Takahiro Shinozaki, Bhiksha Raj, et~al.
\newblock {FreeMatch}: Self-adaptive thresholding for semi-supervised learning.
\newblock In \emph{ICLR}, 2023.

\bibitem[Weiss et~al.(2020)Weiss, Jacob, and Duveiller]{WEISS2020111402}
M. Weiss, F. Jacob, and G. Duveiller.
\newblock Remote sensing for agricultural applications: A meta-review.
\newblock \emph{Remote Sensing of Environment}, 236:\penalty0 111402, 2020.

\bibitem[Xia et~al.(2025)Xia, Niu, and Zhan]{xia2025hierarchical}
Chengwei Xia, Chaoxi Niu, and Kun Zhan.
\newblock Hierarchical consensus network for multiview feature learning.
\newblock pages 21617--21625, 2025.

\bibitem[Xie et~al.(2020{\natexlab{a}})Xie, Dai, Hovy, Luong, and
  Le]{NEURIPS2020_44feb009}
Qizhe Xie, Zihang Dai, Eduard Hovy, Thang Luong, and Quoc Le.
\newblock Unsupervised data augmentation for consistency training.
\newblock In \emph{NeurIPS}, pages 6256--6268, 2020{\natexlab{a}}.

\bibitem[Xie et~al.(2020{\natexlab{b}})Xie, Luong, Hovy, and Le]{Xie_2020_CVPR}
Qizhe Xie, Minh-Thang Luong, Eduard Hovy, and Quoc~V. Le.
\newblock Self-training with noisy student improves imagenet classification.
\newblock In \emph{CVPR}, 2020{\natexlab{b}}.

\bibitem[Xu et~al.(2024)Xu, Zhang, Zhang, and Zhan]{xu2024structure}
Shuaike Xu, Xiaolin Zhang, Peng Zhang, and Kun Zhan.
\newblock Structure-aware consensus network on graphs with few labeled nodes.
\newblock \emph{arXiv:2407.02188}, 2024.

\bibitem[Xue et~al.(2025)Xue, Zhao, Li, Chen, and Zhan]{xueJARS2025}
Tianxiang Xue, Jiayi Zhao, Jingsheng Li, Changlu Chen, and Kun Zhan.
\newblock {CD-Mamba}: Cloud detection with long-range spatial dependency
  modeling.
\newblock \emph{Journal of Applied Remote Sensing}, 19\penalty0 (3):\penalty0
  038507, 2025.

\bibitem[Yang et~al.(2019)Yang, Guo, Yue, Liu, Hu, and Li]{cdnet}
Jingyu Yang, Jianhua Guo, Huanjing Yue, Zhiheng Liu, Haofeng Hu, and Kun Li.
\newblock {CDnet}: {CNN}-based cloud detection for remote sensing imagery.
\newblock \emph{IEEE Transactions on Geoscience and Remote Sensing},
  57\penalty0 (8):\penalty0 6195--6211, 2019.

\bibitem[Yang et~al.(2023)Yang, Qi, Feng, Zhang, and Shi]{Yang_2023_CVPR}
Lihe Yang, Lei Qi, Litong Feng, Wayne Zhang, and Yinghuan Shi.
\newblock Revisiting weak-to-strong consistency in semi-supervised semantic
  segmentation.
\newblock In \emph{CVPR}, pages 7236--7246, 2023.

\bibitem[Yang et~al.(2025)Yang, Zhao, and Zhao]{10839097}
Lihe Yang, Zhen Zhao, and Hengshuang Zhao.
\newblock {UniMatch v2}: Pushing the limit of semi-supervised semantic
  segmentation.
\newblock \emph{IEEE Transactions on Pattern Analysis and Machine
  Intelligence}, 47\penalty0 (4):\penalty0 3031--3048, 2025.

\bibitem[Yao et~al.(2023)Yao, Guo, and Li]{10155420}
Xudong Yao, Qing Guo, and An Li.
\newblock Cloud detection in optical remote sensing images with deep
  semi-supervised and active learning.
\newblock \emph{IEEE Geoscience and Remote Sensing Letters}, 20:\penalty0
  3287537, 2023.

\bibitem[Yun et~al.(2019)Yun, Han, Oh, Chun, Choe, and Yoo]{cutmix}
Sangdoo Yun, Dongyoon Han, Seong~Joon Oh, Sanghyuk Chun, Junsuk Choe, and
  Youngjoon Yoo.
\newblock {CutMix}: Regularization strategy to train strong classifiers with
  localizable features.
\newblock In \emph{ICCV}, pages 6023--6032, 2019.

\bibitem[Zhang et~al.(2018)Zhang, Cisse, Dauphin, and Lopez-Paz]{mixup}
Hongyi Zhang, Moustapha Cisse, Yann~N. Dauphin, and David Lopez-Paz.
\newblock mixup: Beyond empirical risk minimization.
\newblock In \emph{ICLR}, 2018.

\end{thebibliography}
